\documentclass[10pt, a4paper]{article}

\usepackage[final]{lrec2026} 
\usepackage{amsfonts} 
\usepackage{amssymb} 
\usepackage{graphicx}
\usepackage{amsmath}
\usepackage{booktabs} 
\usepackage{hyperref}

\usepackage{dblfloatfix}
\usepackage{caption}

\title{Fine-Grained Perspectives: Modeling Explanations with Annotator-Specific Rationales}

\name{Olufunke O. Sarumi$^{1}$, Charles Welch$^{2}$, Daniel Braun$^{1}$} 

\address{$^{1}$Marburg University, Marburg, Germany\\
         $^{2}$McMaster University, Hamilton, Ontario, Canada \\
         sarumio,daniel.braun@uni-marburg.de, cwelch@mcmaster.ca\\
         }

\abstract{
Beyond exploring disaggregated labels for modeling perspectives, annotator rationales provide fine-grained signals of individual perspectives. In this work, we propose a framework for jointly modeling annotator-specific label prediction and corresponding explanations, fine-tuned on the annotators’ provided rationales. Using a dataset with disaggregated natural language inference (NLI) annotations and annotator-provided explanations, we condition predictions on both annotator identity and demographic metadata through a representation-level User Passport mechanism. We further introduce two explainer architectures: a post-hoc prompt-based explainer and a prefixed bridge explainer that transfers annotator-conditioned classifier representations directly into a generative model. This design enables explanation generation aligned with individual annotator perspectives. Our results show that incorporating explanation modeling substantially improves predictive performance over a baseline annotator-aware classifier, with the prefixed bridge approach achieving more stable label alignment and higher semantic consistency, while the post-hoc approach yields stronger lexical similarity. These findings indicate that modeling explanations as expressions of fine-grained perspective provides a richer and more faithful representation of disagreement. The proposed approaches advance perspectivist modeling by integrating annotator-specific rationales into both predictive and generative components.
 \\ \newline  }

\begin{document}

\maketitleabstract

\section{Introduction}
Perspectivist NLP argues that annotations should reflect the specific judgments of individual annotators rather than converge on a single consensus label \citep{pavlick-kwiatkowski-2019-inherent}. In tasks such as natural language inference (NLI), stance detection, and hate speech classification \citep{xu-etal-2024-leveraging}, it is legitimate for annotators to disagree due to differences in background, interpretation, or socio-demographic perspective. Modeling such disagreement has become an important focus of recent shared tasks \citep{leonardelli-etal-2023-semeval, uma-etal-2021-semeval} and research initiatives, shifting the emphasis away from majority voting toward preserving variation.

Most perspectivist approaches implicitly model perspectives using linguistic and contextual signals such as sociodemographic information, user IDs, and group affiliations \citep{davani2022dealing,plepi-etal-2022-unifying}. These signals are used to infer the potential sources of variation and diversity in annotations. However, beyond disaggregated labels, annotators’ perspectives often remain abstracted and only indirectly represented in the model. Although some datasets require annotators to provide rationales for selecting a particular label \citelanguageresource{weber-genzel-etal-2024-varierr}, these explanations are rarely integrated explicitly into perspectivist modeling. Incorporating annotator rationales enables more nuanced and fine-grained representations of perspective.

Explainability in perspectivist approaches has gradually emerged as a critical component of trustworthy NLP systems, as it supports model-level interpretability through the analysis of attention patterns or internal structures used to justify predictions \citep{mastromattei2022syntax}. In recommendation systems, for example, natural language generation (NLG) methods have been proposed to generate flexible, free-text explanations based on user-generated content \citep{li-etal-2021-personalized}. While such approaches demonstrate the potential of generative models to produce fluent and varied explanations, they also expose limitations: generated content may be off-topic, insufficiently grounded in the input, repetitive \citet{li-etal-2021-implicit}, or insufficiently personalized. These challenges highlight the need for controllable and faithful explanation generation, particularly when explanations are expected to reflect specific user or annotator viewpoints.

Within perspectivist NLP, explainability has been approached in different ways. Some studies treat it as post-hoc model interpretation \citet{mastromattei2022syntax}, identifying linguistic features or structural patterns that influence perspective-aware predictions \citep{Muscato_2025}. Others rely on prompting strategies to simulate user perspectives in large language models \citep{hayati-etal-2024-far}. However, relatively little work has explicitly modeled annotator-specific explanations alongside disaggregated labels, partly because few datasets contain both disagreement and distinct rationales.

In this study, we integrate perspectivist modeling with perspectivist explanation by explicitly conditioning explanation generation on annotator-specific representations. Using a dataset with disaggregated NLI labels and annotator-provided explanations, we model perspective in both label prediction and rationale generation. We explore a prompt-based post-hoc explainer and a representation-prefix bridge that transfers classifier representations enriched with annotator information into a generative model. In doing so, we treat explanations as expressions of perspective rather than merely post-hoc justifications of a model’s decisions.

\section{Related works}
Perspectivist NLP aims to preserve the nuanced information hidden within disagreement by modeling annotator-specific labels rather than aggregating them into a single label \citep{Cabitza_2023}. However, explainability within this paradigm remains relatively understudied and fragmented \citep{frenda2025perspectivist}. Existing research primarily approaches explainability either through model interpretability as in \citet{mastromattei2022syntax} or by explicitly prompting Large Language Models (LLMs) for explanations \citep{orlikowski2025demographicsfinetuninglargelanguage}. However, most work has yet to explore annotator-specific rationales grounded in internal representations as a primary approach for perspectivist explainability.

\subsection{Current Approaches to Perspective-Aware Explanations} One line of research addresses explainability in perspectivist models by identifying the linguistic components in Hate speech tasks with the use of recognizers that incorporates syntactic dependency trees to provide post-hoc justifications for classifications \citep{mastromattei2022syntax}. In these instances, explainability focuses on revealing the mechanics of the model's prediction rather than capturing the annotator’s subjective reasoning.
Similarly, \citet{mastromattei-etal-2022-change} explored explainable syntax-based models within hate speech detection to identify trigger words that influence target classification. In a different vein, \citet{nirmal-etal-2024-towards} implicitly extracted user rationales from input text using LLMs to guide classifier outcomes, aiming for a more interpretable architectural framework.

\subsection{Personalized Generation and Recommendation} A shift toward personalized explanation is evident in the work of \citet{li-etal-2021-personalized}, who designed a specialized Transformer for explainable recommendation. This model utilizes user IDs and items alongside linguistic cues to generate recommendations and justifications that reflect individual user interests. Similarly, \citet{li2020generate} utilized a neural template approach to address user ratings within recommender systems.
More recently, \citet{plepi-etal-2024-perspective} introduced twin-encoder architectures that separately encode auxiliary user information to facilitate perspective-taking in conflict situations. This allows the model to conceptualize user viewpoints through self-disclosure statements. While this approach structurally integrates user context, it does not explicitly disentangle annotator-specific explanatory reasoning in disaggregated datasets, where annotators might agree on a label but diverge significantly in their underlying logic.
In this study, we address explainability through the lens of annotator rationales, seeking to understand the \textit{why} behind a label from the human's perspective. Our approach models annotator perspectives at both the classification and explanation levels. Furthermore, we introduce a representation-level bridge that conditions explanation generation directly on annotator-specific internal representations. By doing so, we treat explanation not merely as a post-hoc interpretability tool, but as an explicit expression of annotators perspectives tied directly to disaggregated labels they represent.

\section{Methods and data}
We study perspectivism in generative explainability using the VariErrNLI dataset \citelanguageresource{weber-genzel-etal-2024-varierr}, which contains disaggregated annotator labels and annotator-specific rationales. Unlike most existing disaggregated datasets, VariErrNLI preserves both label disagreement and explanation diversity, making it suitable for modeling fine-grained perspectives.

Our framework consists of two components: (i) an annotator-aware classifier that predicts label sets for each annotator, and (ii) an annotator-conditioned explainer that generates corresponding rationales. We explicitly model annotator identity using learned embeddings and metadata features, which are fused with the contextual representation of the input (context and statement) to produce annotator-specific predictions.

 We compare two explanation approaches. The first is a post-hoc, prompt-based explainer that generates explanations from textual inputs. The second is a prefixed bridge explainer that conditions generation on the classifier's internal annotator-specific representations. This allows the model to incorporate both predicted labels and underlying annotator-specific reasoning signals.

\begin{table*}[t]
\centering
\small
\setlength{\tabcolsep}{6pt}
\begin{tabular}{lrrrr}
\toprule
\textbf{Statistic} & \textbf{Train} & \textbf{Dev} & \textbf{Test} & \textbf{Total} \\
\midrule
\multicolumn{5}{l}{\textbf{Split-level statistics}} \\
\addlinespace[2pt]
Instances & 388 & 50 & 50 & 488 \\
Annotators & 4 & 4 & 4 & 4 \\
Annotations & 1,505 & 187 & 199 & 1,891 \\
Avg. annotations / instance & 3.88 & 3.74 & 3.98 & 3.88 \\
Explanations & 1,505 & 187 & 199 & 1,891 \\
Avg. explanation length (words) & 13.90 & 13.12 & 14.28 & 13.86 \\
\addlinespace[4pt]
\multicolumn{5}{l}{\textbf{Label distribution (count, \%)}} \\
Entailment & 446 (29.6\%) & 34 (18.2\%) & 61 (30.7\%) & 541 (28.6\%) \\
Neutral & 767 (51.0\%) & 96 (51.3\%) & 93 (46.7\%) & 956 (50.6\%) \\
Contradiction & 292 (19.4\%) & 57 (30.5\%) & 45 (22.6\%) & 394 (20.8\%) \\
\addlinespace[4pt]
\multicolumn{5}{l}{\textbf{Annotations per annotator (count) and demographics}} \\
Ann1 (F,22,CN,MSc) & 367 & 45 & 47 & 459 \\
Ann2 (M,33,DE,Postdoc) & 376 & 45 & 47 & 468 \\
Ann3 (F,25,CN,MSc) & 379 & 46 & 54 & 479 \\
Ann4 (M,25,CN,MSc) & 383 & 51 & 51 & 485 \\
\bottomrule
\end{tabular}
\caption{VariErrNLI dataset statistics by split. Demographics are abbreviated as Gender, Age, Nationality, Education (CN=Chinese, DE=German; MSc=Master student).}
\label{tab:VariErrNLI_stats}
\end{table*}

\subsection{VariErrNLI Dataset}
We use VariErrNLI (Variation vs. Error), a perspectivist NLI dataset designed to disentangle human label variation from annotation error. VariErrNLI contains approximately 500 NLI items sampled from ChaosNLI (MNLI subset) and annotated in two rounds by four independent annotators.

 In Round 1, annotators assigned one or more NLI labels, Entailment (E), Neutral (N), or Contradiction (C) to each item and provided a one-sentence explanation for each label assigned, preserving fine-grained reasoning diversity. This round of annotation produced 1,933 label-explanation pairs.

 In Round 2, annotators independently evaluated the validity of each label–explanation pair (including their own) by judging whether the explanation plausibly supports the assigned label. This second stage enables distinguishing plausible human label variation from annotation errors. The dataset, therefore, provides not only disaggregated labels and rationales but also meta-judgments about their validity. 

 Although VariErrNLI was originally designed to study annotation error versus variation, we use it for a different purpose. Specifically, we leverage its disaggregated labels and annotator-specific explanations to model and generate annotator-conditioned reasoning. For this study, we use the version released for the Learning with Disagreement (LeWiDi) 2025 Shared Task \citelanguageresource{leonardelli2026lewidi2025nlperspectiveseditionlearning}, which provides predefined training, development, and test splits. The dataset statistics are presented in Table~\ref{tab:VariErrNLI_stats} 

\subsection{Problem Formulation}
We formalize annotator-specific prediction and explanation as a joint task. Each instance in the VariErrNLI dataset consists of a context $c$, a statement $s$, and annotations from annotators $a \in \mathcal{A}$. Each annotator provides a judgment (in some instances, multi-label)  over the label set
    \begin{equation}
        \mathcal{L} = \{C, E, N\},
    \end{equation}
    corresponding to contradiction (C), entailment (E), and neutral (N); and an explanation that justifies their labeling decision. For each annotator $a$, there is an annotator-specific label 
\begin{equation}
    y_a \subseteq \mathcal{L},
\end{equation}
and an associated explanation $r_a$, where $r_a$ is a short sentence describing the reasoning for $y_a$. Because two annotators can assign the same label for different reasons, we treat explanation generation as an explicitly perspectivist problem. Our model therefore has two goals: (i) predict the annotator-specific label set for each annotator $a$, and (ii) generate the corresponding annotator explanation $ r_a$, which we define as the annotator's expressed perspective. For each instance $(c,s)$ and annotator $a$, we learn an annotator-aware classifier and an annotator-conditioned explainer trained on the provided human rationales.  

\subsection{Annotator-Aware Classification}
We implement the \textit{User Passport} method to explicitly model annotator-specific perspectives within our classification framework \citep{sarumi-etal-2025-nlp}, using DeBERTa-v3-base as the backbone encoder. This approach incorporates annotator identity and metadata directly at the representation level rather than through input text modification or token-based methods \citep{welch-etal-2022-leveraging}. The resulting classifier serves as the underlying prediction component for both the post-hoc and prefixed bridge explanation models. 

Formally, we consider an annotated dataset defined by $\mathcal{D} = (X, A, Y)$, where $X$ is the set of text instances $\{x_1, x_2, \dots, x_n\}$. Each instance $x_i \in \mathcal{X}$ is a pair $(c_i, s_i)$ representing the context and statement. The set $A = \{a_1, a_2, \dots, a_k\}$ represents unique annotators, and the annotation matrix is defined as:
\begin{equation}
    Y : X \times A \rightarrow \{0,1\}^{3}
\end{equation}
To handle varying annotator coverage, a masking mechanism is applied during training and evaluation. The annotator-level loss is computed only for instances where a label exists, using a binary mask to ensure missing annotations do not contribute to the training objective.

The encoder extracts a pooled representation $h \in \mathbb{R}^{H}$ capturing the relationship between $c_i$ and $s_i$. To incorporate individual variation, we define a learnable embedding space where each annotator $a_j$ is mapped to a unique, $d$-dimensional vector $u_j \in \mathbb{R}^{E}$:
\begin{equation}
    u_j = \mathrm{Embedding}(a_j)
\end{equation}
Simultaneously, each annotator’s structured demographic metadata is transformed into a fixed-size vector $m_j$ and projected into the latent space of the text encoder. We then perform a \textit{representation-level fusion} by concatenating the instance representation, the annotator embedding, and the metadata projection. The resulting fused representation $z_{ij}$ is passed to the classification head:
\begin{equation}
    z_{ij} = [\, h \,;\, u_j \,;\, m_j]
\end{equation}
This allows the model to explicitly account for both the annotator’s identity and their demographic context by learning systematic patterns between these features and labeling behavior through latent feature fusion.

\subsection{Annotator Explanation Modeling}
To generate annotator-specific rationales, we implement two explanation approaches that produce an explanation  $r_a$ but differ in how they incorporate classifier information. 

\subsubsection{Post-hoc Explainer}
Our first approach trains a standard encoder-decoder model \(Flan-T5\) \citelanguageresource{10.5555/3722577.3722647} to generate an annotator explanation using a text-only prompt. For each training record, we construct an input prompt that contains: the context and statement, the annotator's gold labels, annotators persona: derived from the annotator metadata information, and an annotator control token. The annotator control token is linked to the annotator ID in the dataset and prepended to the prompt by extending the tokenizer vocabulary with a unique, learnable special token \citep{sarumi-etal-2024-corpus,plepi-etal-2022-unifying}. At inference time, we insert the classifier's predicted probabilities $(p_{C}, p_{E}, p_{N})$ into the prompt. The explainer then generates a short explanation. In this setup, there is no differentiable connection between the classifier and the explainer.

\subsubsection{Prefixed Bridge Explainer}
\begin{figure}[t]
    \centering
    \includegraphics[width=\columnwidth]{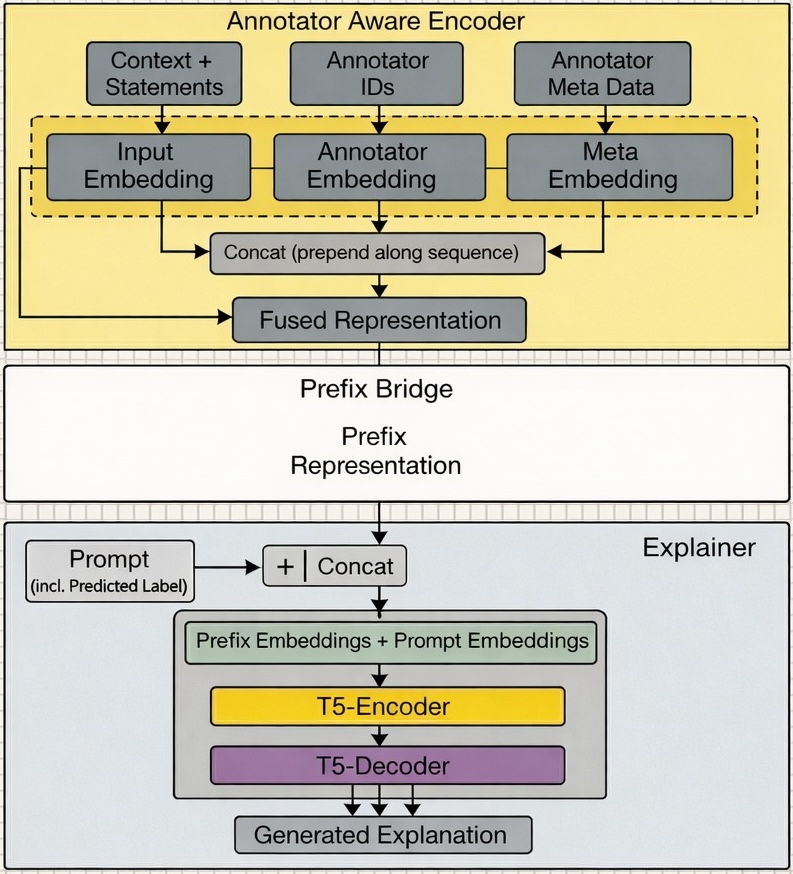}
    \caption{The Prefixed Bridged Explainer}
    \label{fig:prefixed_explainer}
\end{figure}

Our second approach introduces a stronger coupling between classification and explanation using the classifier’s continuous internal representation, rather than text-only features.
We first run the annotator-aware classifier on $(c, s)$ to obtain the fused representation $z_{ij}$. We then learn a small neural Prefixed Bridge (a 2-layer MLP) that projects this vector into a sequence of prefix embeddings with the same dimensionality as the T5 encoder embedding space. These prefix embeddings are prepended to the T5 encoder input embeddings before encoding. We then
train by freezing the classifier parameters and optimizing the bridge and the T5 parameters to minimize explanation generation loss. At inference time, explanation generation is performed using the prefix produced by the bridge, which is concatenated with the prompt token embeddings before encoding and generation (see Figure~\ref{fig:prefixed_explainer}).

\begin{table*}[!t]
\centering
\small
\begin{tabular*}{\textwidth}{@{\extracolsep{\fill}} l c c c c @{}}
\hline
\textbf{Explainer} & \textbf{F1 (Macro)} & \textbf{Exact Match} & \textbf{ROUGE-L} & \textbf{Semantic Similarity} \\
\hline
User Passport \citep{sarumi-etal-2025-nlp} & 70.5 & —  & —  & — \\
Post-hoc Explainer & 92.3 & 92.2 & \textbf{24.5} & 51.0 \\
Prefixed Bridge Explainer & \textbf{93.9} & \textbf{92.4} & 24.0 & \textbf{53.4}\\
\hline
\end{tabular*}

\vspace{2mm}
\caption{\fontsize{10}{12}\selectfont Aggregated evaluation scores across all annotators. We report the results of the User Passport model from previous work, without explanation, as the baseline. Bold values indicate the best scores. All scores are reported as the mean of three runs.}
\label{tab:aggregated_scores}
\end{table*}

\section{Experiments}
In our experiments, we used two base models that follow the encoder-decoder architecture. We also implemented the User Passport method for incorporating annotator-meta information.

\subsection{Experimental set-up}
We train the annotator-aware classifier for 50 epochs using the AdamW optimizer with a learning rate of $2 \times 10^{-5}$ and weight decay $0.01$. A linear scheduler with warmup (ratio $0.06$),  gradient clipping (max norm $1.0$), and early stopping on development macro-F1 (patience 3) is applied. The backbone model is DeBERTa-v3-base \citelanguageresource{he2023debertav3improvingdebertausing}, with a maximum input length of 256 and batch size 32. To model annotator-specific predictions, we incorporate annotator information through a learnable annotator embedding (dimension 64) and a projected metadata representation, fused with the instance representation at the feature level. Training uses masked binary cross-entropy with an auxiliary soft-label alignment objective ( $\lambda_{\mathrm{soft}} = 1.0$). Class imbalance is handled using masked focal BCE with class-specific positive weighting.

 For explanation generation, both explainer variants are trained using Flan-T5-base with a maximum input length 512 and target length 128. Models are trained for up to 50 epochs with early stopping on validation loss, using AdamW with learning rate $8 \times 10^{-5}$ and weight decay $0.01$. Label thresholds are tuned on the development set with grid search over $[0.1, 0.9]$, selecting the configuration that maximizes mean Jaccard similarity with gold annotator label-sets.

 All experiments are conducted on a single NVIDIA A100 80GB PCIe GPU (CUDA 13.1). Average end-to-end runtime (training and evaluation) is approximately 15-20 minutes per model. All reported results are averaged over three runs.

\begin{table*}[t]
\centering
\small
\begin{tabular}{l l l l l c c c c}
\hline
\multicolumn{9}{c}{\textbf{Prefixed Bridge Explainer}} \\
\hline
Annotator & Gender & Age & Nationality & Education & Macro F1 & Exact Match & ROUGE-L & Semantic Sim \\
\hline
Ann1 & Female & 22 & Chinese & MSc. & \textbf{94.3} & \textbf{94.2} & 23.8 & 55.2 \\
Ann2 & Male & 33 & German & Postdoc & 92.5 & 92.0 & \textbf{34.1} & \textbf{59.6} \\
Ann3 & Female & 25 & Chinese & MSc & 92.0 & 88.7 & 21.2 & 53.5 \\
Ann4 & Male & 25 & Chinese & MSc & \textbf{95.5} & \textbf{94.7} & 17.9 & 45.8 \\
\hline
\multicolumn{9}{c}{\textbf{Post-hoc Explainer}} \\
\hline
Annotator & Gender & Age & Nationality & Education & Macro F1 & Exact Match & ROUGE-L & Semantic Sim \\
\hline
Ann1 & Female & 22 & Chinese & MSc. & 87.9 & 90.6 & 24.5 & 49.9 \\
Ann2 & Male & 33 & German & Postdoc & \textbf{96.7} & \textbf{97.1} & \textbf{31.3} & \textbf{55.9} \\
Ann3 & Female & 25 & Chinese & MSc & 90.4 & 88.7 & 22.9 & 50.0 \\
Ann4 & Male & 25 & Chinese & MSc & 92.4 & 92.7 & 19.6 & 48.7 \\
\hline
\end{tabular}
\caption{\fontsize{10}{12}\selectfont Comparison of Explainers per annotator: A descriptive Analysis. Bold values highlight key patterns discussed in section 5: improved predictive performance with the bridge model (Ann1, Ann4), stronger lexical overlap and semantic strength with both prefixed and post-hoc model (Ann2), and overall Macro-F1 score, Exact Match (notably Ann2).}
\label{tab:combined_explainer_results}
\end{table*}




\begin{figure}[t]
    \centering
    \includegraphics[width=\columnwidth]{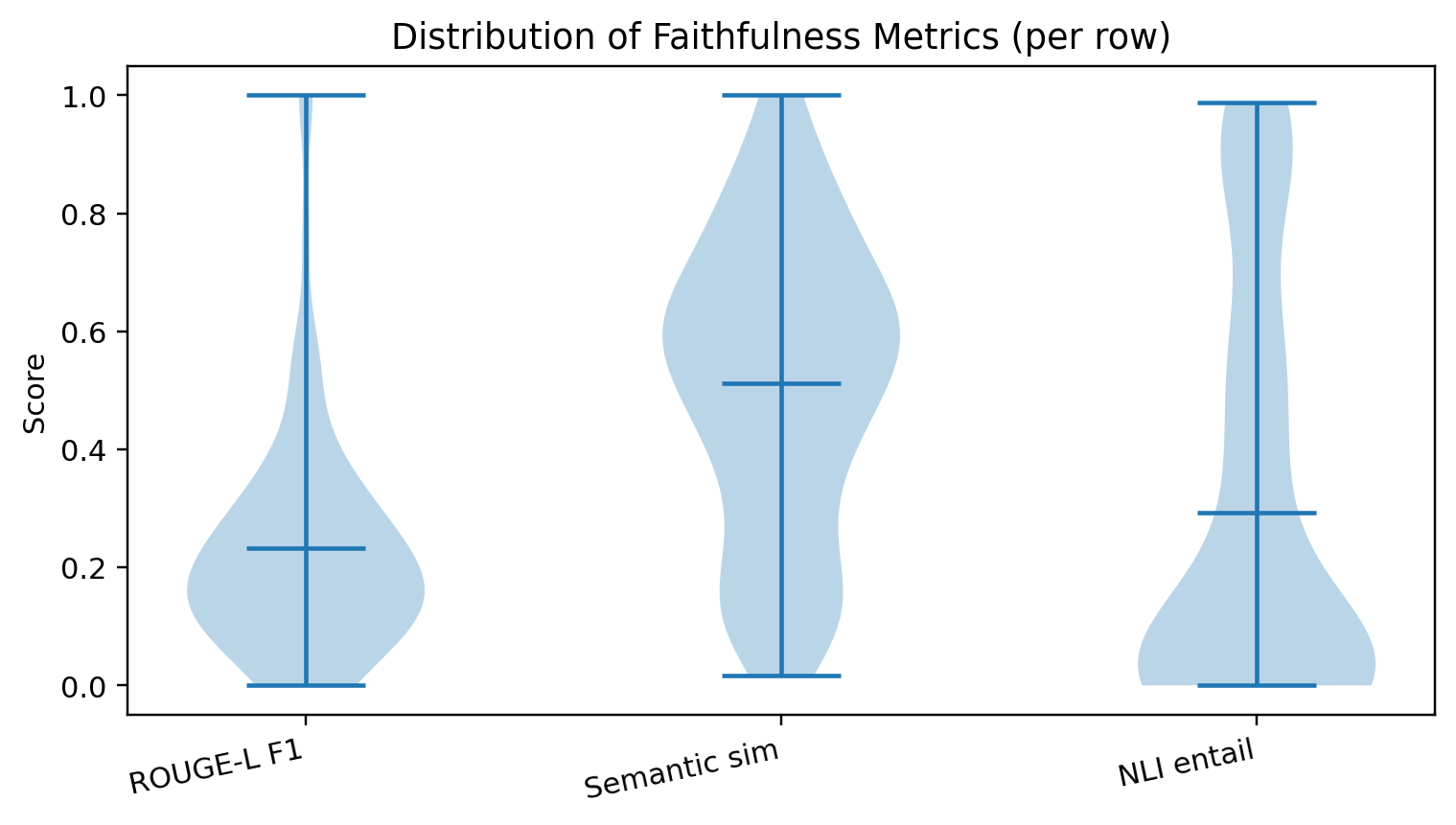}
    \caption{Prefixed Bridged Faithfulness Evaluation}
    \label{fig:prefixed_distribution}
\end{figure}

\begin{figure}[t]
    \centering
    \includegraphics[width=\columnwidth]{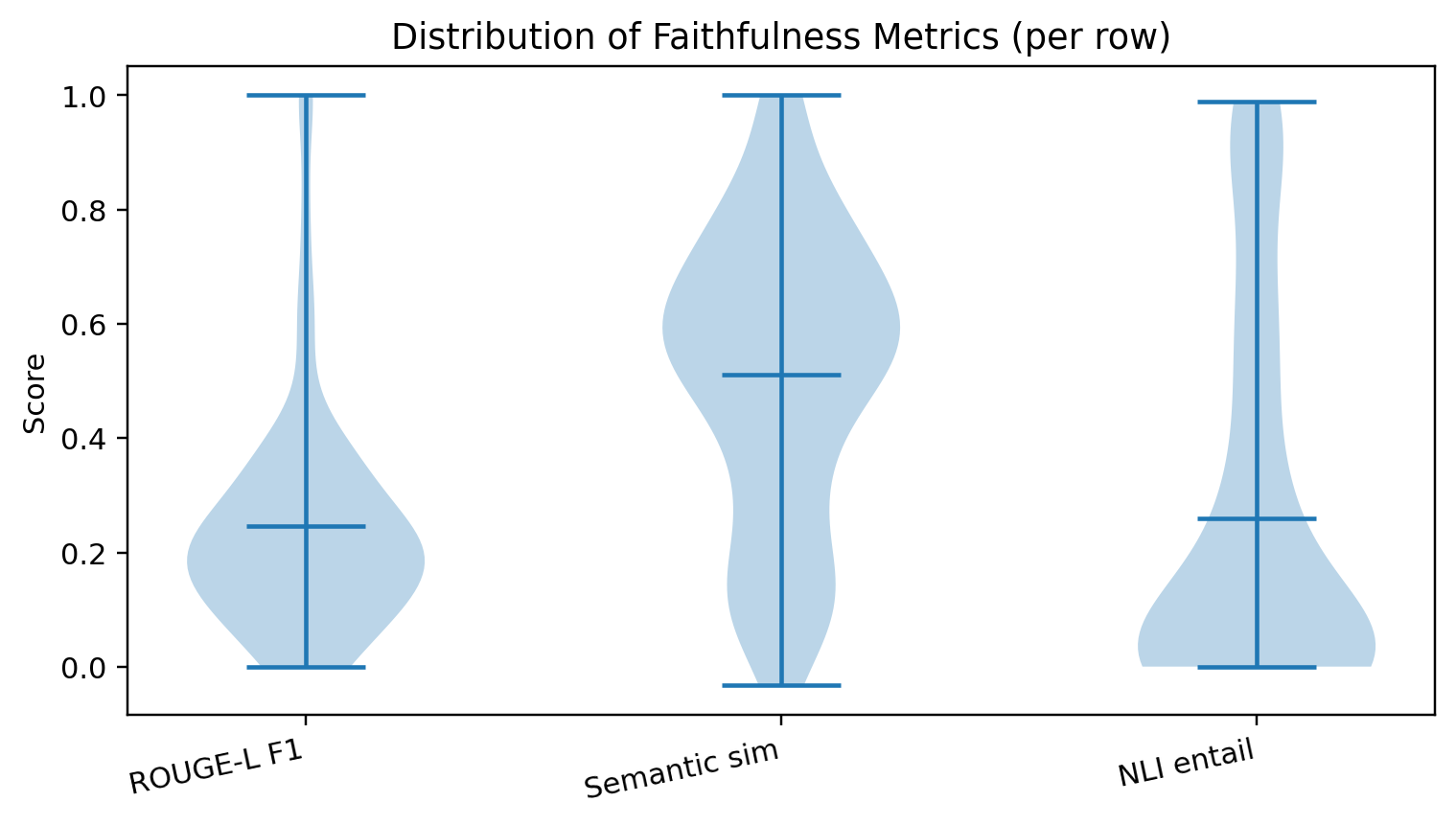}
    \caption{Post-hoc Faithfulness Evaluation}
    \label{fig:posthoc_distribution}
\end{figure}

\section{Result and Discussion}
\subsubsection*{Aggregated Evaluation of Explainers}Table~\ref{tab:aggregated_scores} presents the aggregated performance comparison between the baseline annotator-aware classifier (User Passport) from previous work, the Post-hoc Explainer, and the Prefixed Bridge Explainer. The baseline achieves a Macro-F1 score of 70.5, indicating that incorporating explanation modeling substantially improves classification performance.

Both explanation-based approaches outperform the baseline, with the Prefixed Bridge Explainer achieving the highest Macro-F1 (93.9) and Exact Match (92.4), indicating stronger agreement with the gold labels. The Post-hoc Explainer also performs well with Macro-F1(92.3), but remains slightly below the bridge model.

In terms of explanation quality, ROUGE-L is marginally higher for the Post-hoc Explainer, suggesting better lexical overlap with reference explanations, which is consistent with its text-based approach. In contrast, the Prefixed Bridge Explainer achieves higher semantic similarity, indicating that its generated explanations are better aligned in meaning. This improvement can be attributed to its use of the classifier’s internal representations, which provide richer contextual features for generation.

These results show that explanation modeling significantly improves performance over the baseline, while tighter integration between prediction and generation further enhances classification consistency and semantic alignment.
\begin{figure*}[t]
    \centering
    \includegraphics[width=\textwidth,height=0.6\textheight,keepaspectratio]{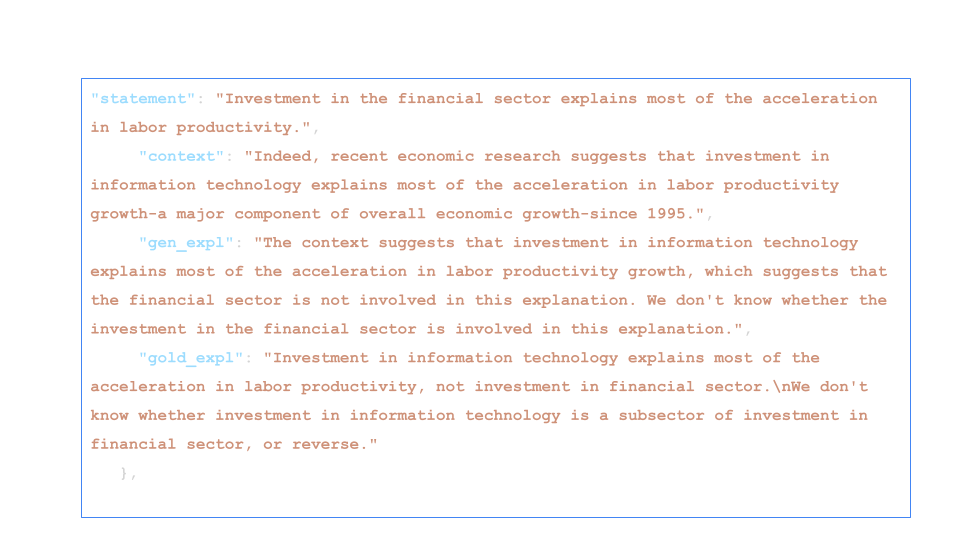}
    \caption{The Prefixed Bridged Explanation Example}
    \label{fig:prefixed_explanation_sample}
\end{figure*}

\begin{figure*}[t]
    \centering
    \includegraphics[width=\textwidth,height=0.6\textheight,keepaspectratio]{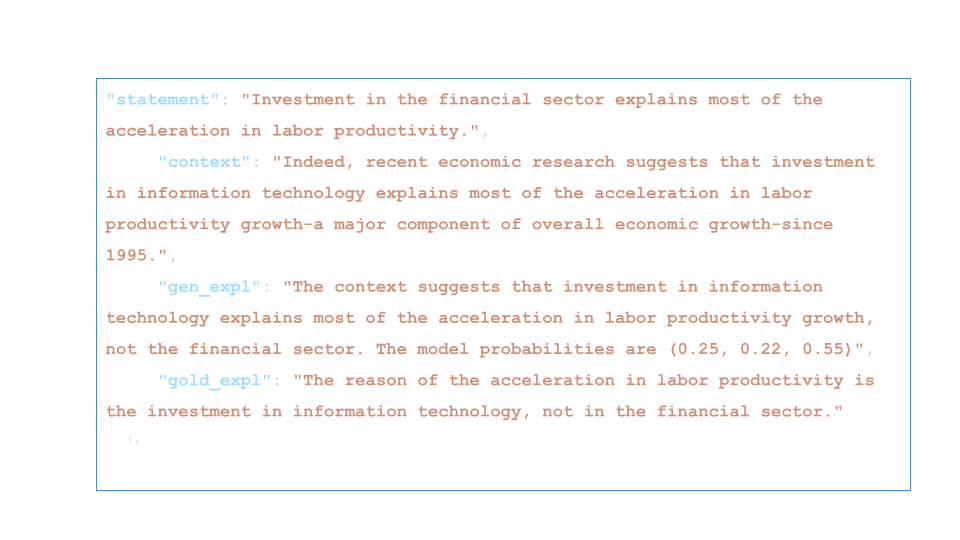}
    \caption{The Post-hoc Explanation Example}
    \label{fig:post-hoc_explanation_sample}
\end{figure*}

\subsubsection*{Faithfulness Distribution and Qualitative Analysis}
The faithfulness distributions in Figures~\ref{fig:prefixed_distribution} and~\ref{fig:posthoc_distribution} show that, for both models, semantic similarity scores cluster around moderate values (median $\sim$ 0.5), while ROUGE-L remains relatively low (median $\sim$ 0.24), indicating lexical divergence despite semantic alignment. However, the Prefixed Bridge Explainer exhibits a more balanced NLI entailment distribution, with a larger proportion of high-entailment cases compared to the Post-hoc model, suggesting stronger inferential alignment between predictions and explanations. 
 To further examine these differences, we present qualitative examples in Figures~\ref{fig:prefixed_explanation_sample} and~\ref{fig:post-hoc_explanation_sample}, focusing on cases where the predicted label is consistent but the generated explanations differ in structure and depth. In both examples, the two models correctly identify that the context supports investment in information technology rather than the financial sector. However, the nature of the generated explanations differs. The Prefixed Bridge Explainer produces explanations that are more concise and directly grounded in the key contrast between the context and the statement, closely mirroring the underlying reasoning required for the prediction. In contrast, the Post-hoc Explainer tends to generate more verbose explanations, introducing additional statements that are not explicitly stated in the context. While both explanations are semantically aligned with the gold rationale, the prefixed bridged explanation more precisely captures the core inference without introducing random words. This qualitative difference supports the distributional findings. The Prefixed Bridge Explainer demonstrates stronger alignment between prediction and explanation, not only quantitatively through higher scores, but also qualitatively in the clarity and focus of the generated explanation.

\subsubsection*{Comparison of Explainers per annotator: A descriptive Analysis.}
Table~\ref{tab:combined_explainer_results} presents a detailed examination across the four annotators. Differences are observed in Macro-F1, Exact Match, ROUGE-L, and Semantic similarity, suggesting that both models interact differently with individual annotator patterns.

The Prefixed Bridge Explainer generally produces more stable performance across annotators in terms of Macro-F1 and Exact Match. In particular, Ann1 and Ann4 show improvements in both metrics compared to the Post-hoc Explainer, indicating that incorporating classifier-level representations contributes to more reliable alignment between predictions and annotator-specific labels. This suggests that the shared representation between the two encoders better captures variability in annotator decision patterns, especially when explanations differ in structure or clarity.

Ann2 achieves the highest overall performance, particularly under the Post-hoc Explainer (Macro-F1: 96.7, Exact Match: 97.1), outperforming the Prefixed Bridge model. Ann3 and Ann4 exhibit comparatively lower or more variable performance across certain metrics, particularly in ROUGE-L. For Ann4, while Macro-F1 and Exact Match improve under the Prefixed Bridge Explainer, ROUGE-L and Semantic similarity remain relatively low across both models.

A closer examination of the VariErrNLI dataset \citep{weber-genzel-etal-2024-varierr} provides important context for interpreting these results. The dataset explicitly distinguishes between variation and annotation error through a second round of self- and peer-validation, where explanations are assessed for whether they plausibly support the assigned labels. As shown in the original study, agreement increases substantially after validation, indicating that a portion of annotator disagreement is attributable not to genuine perspectives and differences, but to inconsistencies and errors.

This distinction is reflected in our findings. Annotators whose explanations are more consistently grounded in the input text and validated by peers are more reliably modeled by both approaches. In particular, Ann2 achieves the highest predictive performance across metrics, especially under the Post-hoc Explainer, aligning with the dataset’s validation framework where more coherent and text-aligned reasoning leads to more stable label–explanation pairs. Notably, Ann2 is also the annotator with the highest age (33) and level of education (Postdoc) in the dataset. While this may be associated with clearer or more structured explanations, stronger task understanding or domain expertise, we do not draw definitive conclusions from this observation due to the limited number of annotators. Instead, this serves as an indicative pattern that can be further investigated in larger and more controlled settings.

In contrast, annotators exhibiting more variability in explanation quality are more challenging to model. For example, Ann4 shows comparatively lower or less consistent performance across certain metrics, particularly in lexical overlap (ROUGE-L), despite improvements in predictive performance under the Prefixed Bridge Explainer. This pattern is consistent with the dataset observations, where some explanations may be less well-aligned with the assigned labels or expressed in ways that deviate from reference formulations. As a result, the model relies more heavily on underlying representations rather than surface-level cues.

A consistent pattern across annotators is the divergence between lexical and semantic metrics. The Post-hoc Explainer tends to produce higher ROUGE-L scores, indicating closer surface-level similarity to reference explanations. In contrast, the Prefixed Bridge Explainer achieves higher or comparable semantic similarity across most annotators, suggesting better alignment in meaning. This reflects the underlying modeling difference: the Post-hoc approach relies primarily on textual prompts, whereas the bridge model leverages classifier-derived internal representations, enabling richer contextual grounding of explanations.

These per-annotator differences highlight that identical labels do not imply identical reasoning processes. The variation observed across metrics suggests that annotators may express similar decisions through different explanatory structures, levels of detail, or linguistic forms. By incorporating explanation generation, both models move beyond label prediction and provide additional insight into how annotator perspectives are represented. The Prefixed Bridge Explainer, in particular, better preserves the relationship between predictions and underlying reasoning, especially when explanations are less consistent in form.

It is important to note that these observations are based on a small number of annotators, with limited demographic diversity and a relatively small test set. As such, we do not perform statistical significance testing and instead rely on descriptive analysis. The patterns observed should therefore be interpreted as indicative trends rather than generalizable findings.

Overall, the per-annotator analysis suggests that incorporating explanation modeling improves the representation of annotator perspectives, and that tighter integration between prediction and explanation, as in the Prefixed Bridge Explainer, provides more consistent and semantically aligned outputs across diverse annotator behaviors.

\section{Conclusion}
This work demonstrates the importance of Modeling fine-grained annotator perspectives jointly with explanation generation in natural language inference. Rather than treating explanations as post-hoc rationalizations, we show that integrating annotator-expressed rationales into the predictive architecture enables more robust modeling of human diversity. By leveraging explanation-level supervision tied to individual annotations, the model captures not only label outcomes but also the reasoning patterns underlying them, allowing for more faithful representation of disagreement and interpretative nuance.

Methodologically, we implement an encoder-to-encoder bridge architecture that explicitly connects prediction and explanation modules. This structural coupling enables the model to condition its explanatory representations on the same signals that drive classification decisions, thereby improving macro-level stability and inferential alignment across annotators. Our results show that Modeling perspectives through annotator rationales strengthens semantic consistency and predictive robustness, particularly in semantically complex categories. Overall, this work highlights the value of integrating explanation modeling into annotator-aware architectures for developing more transparent and perspective-sensitive NLP systems.

\section{Limitation}
A primary limitation of this work is the dataset used, which was originally constructed to investigate annotation errors in human label variation. Although the inclusion of annotator-specific rationales represents a substantial step toward preserving individual reasoning patterns, the explanations were not designed to systematically capture controlled variations in demographic, linguistic, or cultural background, but were instead targeted toward annotation error detection. As a result, the scope remains limited, which may constrain the generalizability of the proposed encoder-to-encoder bridge framework.

We initially proposed extending the ChaOSNLI instances used in VariErrNLI with explanations written by native English speakers to systematically examine how bilingual versus native annotator rationales affect modeling outcomes. This extension would allow a more controlled investigation of linguistic background effects on explanation faithfulness and predictive performance. Future work will focus on expanding the dataset in this direction to strengthen the empirical foundation of perspective modeling. 

Additionally, an ensemble of the Post-hoc and Prefixed Bridge approaches presents an interesting direction for future work, as it could leverage the strengths of the individual models to produce a more well-rounded output. 

All code and resources developed for this study are publicly available\footnote{\href{https://github.com/Responsible-NLP/LRECNLPerspectives2026-Fine-Grained-Perspective}{https://github.com/Responsible-NLP/LRECNLPerspectives2026-Fine-Grained-Perspective}} to facilitate reproducibility and further research.

\section*{Bibliographical References}\label{sec:reference}
\bibliographystyle{lrec2026-natbib}
\bibliography{references}

\section*{Language Resource References}
\label{lr:ref}
\bibliographystylelanguageresource{lrec2026-natbib}
\bibliographylanguageresource{languageresource}

\end{document}